# Multitask Conditional Generative Adversarial Network Enables Automatic Whole Knee Cartilage and Menisci Segmentation and Reliable T1ρ and T2 Quantification Without High-Resolution Morphological Images

Ahmed Tahseen Minhaz[1,2], Richard Lartey[1,2], Zhiyuan Zhang[1,2], Jeehun Kim[1,2], Kunio Nakamura[1,2], Mingrui Yang[1,2], Jiasen Zhang[1,3], Weihong Guo[3], Naveen Subhas[1,4], Carl S. Winalski[1,2,4], Xiaojuan Li[1,2,4]

***Abstract*— Early osteoarthritis detection through quantitative MRI (qMRI) requires accurate cartilage and meniscus segmentation, traditionally necessitating time-consuming, costly 3D high-resolution Double Echo Steady-State (DESS) MRI scans. This study developed a multi-task conditional generative adversarial network (MT-cGAN) to simultaneously synthesize DESS-like images and segment tissues directly from qMRI echo images. This retrospective study evaluated 508 knee MRI volumes from 361 subjects (mean age: 40.4 ± 12.2 years; 179 female) across three cohorts. Ground truth segmentation masks were generated from DESS images using a pretrained model with manual correction, and T1ρ and T2 maps were computed from magnetization-prepared angle-modulated partitioned k-space spoiled gradient echo snapshots (MAPSS) echo images. MT-cGAN was trained to jointly synthesize DESS-like images and segment cartilage and meniscus directly from echo images. Model performance was evaluated using Dice score for segmentation accuracy and coefficient of variation (CV) for T1ρ and T2 quantification. MT-cGAN achieved the highest segmentation performance, mean Dice score 0.84 (range: 0.80–0.86) across all cartilage and meniscus compartments and significantly outperformed the state-of-the-art conditional GAN model with transfer learning (mean Dice, 0.82; $p < 0.001$, Wilcoxon signed-rank test). For relaxometry quantification, MT-cGAN demonstrated the highest consistency with the reference DESS protocol, yielding the lowest CV (T1ρ: 1.84%, T2: 1.81%). Proposed MT-cGAN accurately segmented cartilage and menisci while providing reliable T1ρ and T2 quantification directly from echo images. By eliminating the need for separate morphological DESS scans, this workflow reduces required scan times to facilitate the clinical translation of qMRI.**

## I. INTRODUCTION

Osteoarthritis (OA) affects over 32.5 million adults in the United States[1,2]. Early detection of cartilage degeneration is crucial to prevent further progression of OA. Quantitative MRI (qMRI) techniques, e.g. $T_{1\rho}$ and $T_2$ mapping can detect proteoglycan loss, collagen disruption, and free water content increase, that occur well before morphological changes in cartilage during degeneration[3–5]. Accurate cartilage and meniscus quantification requires precise segmentation of these structures. In clinical and research practice, segmentation is often performed on high-resolution morphological MR images such as 3D Double Echo Steady-State (DESS) to delineate cartilage and meniscus, and those masks are then overlaid on T1ρ and T2 maps to extract relaxometry values[6,7]. However, acquiring an additional 3D-DESS for segmentation increases total scan time by approximately 6 minutes and thus cost. Furthermore, image registration is needed when patient motion occurs between DESS and qMRI image acquisitions. Therefore, eliminating the need for DESS by segmenting directly from the echo images acquired during qMRI would streamline protocols, reduce scan time, avoid potential registration errors, and expand the clinical feasibility of compositional knee MRI.

Automatic segmentation of knee cartilage from knee MR images is an active area of research. Traditional methods employed atlas-based techniques[8–12] or statistical shape models[13–17], but these required laborious tuning and often struggled with accuracy and generalizability. In recent years, deep learning (DL) has revolutionized medical image segmentation, including musculoskeletal applications[18–39]. Prasoon et al. demonstrated one of the first DL models for tibial cartilage segmentation using a triplanar convolutional neural network (CNN)[18]. Liu et al. developed a SegNet-based CNN to segment cartilage and bone from multi-contrast knee MRI[19]. The model was trained and evaluated on 3D-SPGR, 3D-Turbo-Spin Echo (TSE), and T2 mapping images. Zhou et al. built on this

[1]Program of Advanced Musculoskeletal Imaging (PAMI), Cleveland Clinic
[2]Department of Biomedical Engineering, Cleveland Clinic Research, Cleveland Clinic
[3]Department of Mathematics, Case Western Reserve University
[4]Department of Diagnostic Radiology, Cleveland Clinic

This work was supported by the National Institutes of Health / National Institute of Arthritis and Musculoskeletal and Skin Diseases (NIH/NIAMS) under grant number R01AR077452, R01AR075422 and K25AR078928.

work with a new encoder-decoder CNN combined with 3D conditional random fields and deformable modeling to segment 12 knee joint structures from 3D-FSE images[20]. Raj et al. proposed a 3D-CNN to segment cartilage and meniscus in high resolution DESS images[21]. Norman et al. developed a 2D U-Net for cartilage and meniscus segmentation from SPGR T1ρ-weighted and 3D-DESS knee MRIs[22]. A hybrid approach combining 3D statistical shape models with CNNs was proposed by Ambellan et al., becoming the first to achieve inter-observer equivalent accuracy on the SKI10 challenge[23]. Conditional GANs were compared against U-Net for multi-tissue knee segmentation, with results showing that a tuned cGAN matched U-Net performance (Dice ≥ 0.95 for bone, ≥ 0.83 for cartilage and muscle) and that transfer learning preserved learned capabilities across datasets[24]. Gaj et al. introduced a conditional generative adversarial network (Unet-cGAN) to segment knee cartilage and meniscus and achieved state-of-the-art performance (mean cartilage Dice 0.88, mean meniscus Dice 0.88) by introducing a modified objective function that combined Dice loss, cross-entropy loss with feature loss[25]. A 3D neural network with adversarial loss was proposed for prior shape-constrained bone segmentation combined with a U-Net for cartilage, showing robustness for pathological cases on the SKI10 benchmark[30]. Several additional studies have further validated automated segmentation across diverse datasets and clinical protocols[26–29,31–36]. Multiple architectures, UNet, UNet++, ResUNet, and TransUNet, were compared on a mixed 700-case dataset for automated cartilage segmentation, thickness measurement and grading, with TransUNet performing the best[37]. Ferreira et al. introduced SAMRI-2, a transformer-based vision foundational model that outperformed 3D nnU-Net and other baselines while requiring as few as three user clicks per volume[38]. Moorthy et al. proposed a unified attention-driven framework, 3D CSFA-UNet to jointly perform knee MRI segmentation and OA severity classification in a single end-to-end architecture[39].

Despite these advances, most prior works assume the availability of a high-resolution, high-contrast image like 3D DESS or other sequences (3D TSE/FSE, 3D SPGR/FLASH) for segmentation. Quantitative MRI sequences, T1ρ and T2 mapping, which are most relevant to compositional assessment of cartilage and meniscus degeneration, are almost never used as the segmentation input itself. Norman et al. used T1ρ-weighted image-based segmentation as an alternative to DESS, motivated by its biochemical relevance. However, performance on the T1ρ dataset was consistently lower across all compartments in validation and test cases, with particularly poor generalization for patellar and medial tibial cartilage. This likely stemmed from the smaller sample size and lower spatial resolution of the T1ρ data[22]. This highlights that while echo images are biochemically rich, they are not traditionally considered optimal for segmentation. To address this, Neubert et al. proposed syn3D-DESS, a synthetic DESS image generation approach from 2D-TSE to improve segmentation. While the image generation was performed using U-net, the automated segmentation was atlas-based[40]. Although this approach improved segmentation accuracy, the two-step process was suboptimal and did not jointly learn both tasks. Moreover, the reliance on separate synthesis and segmentation models can introduce compounding errors and require more computational steps.

An emerging solution to such limitations is multi-task learning, where a single model learns multiple objectives simultaneously[41]. This training strategy encourages shared feature learning across tasks and has been shown to improve overall model generalization[42,43]. Multitask learning has been implemented in medical image related tasks[44,45]. Karkalousos et al. demonstrated that a multitask model that combines MRI reconstruction and brain lesion segmentation outperformed separate models for each task, improving both image quality and segmentation[46].

In this study, we propose an end-to-end multitask learning framework, multi-task conditional generative adversarial network (MT-cGAN), that jointly segments cartilage and meniscus while synthesizing a DESS-like image from quantitative echo images. Unlike prior studies, our method operates directly on MAPSS T1ρ and T2 mapping echo images already acquired for quantitative mapping. To our knowledge, this is the first work to demonstrate that high-quality segmentation of cartilage and meniscus is possible directly from echo images used for quantitative mapping, enabling one-stop imaging for both knee cartilage/meniscus morphology and composition.

## II. METHODS

### A. Dataset

Table 1a shows the three independent IRB approved studies that contributed to the dataset used for model development, validation, and testing. Across all studies, DESS and T1ρ and T2 mapping MRI sequences were utilized for quantitative morphological assessment of knee cartilage. MRI of all studies were collected on 3T MR scanners with dedicated knee coils (Table 1b). Major parameters from the MRI imaging protocol are summarized in Table 1c.

The T1ρ and T2 imaging was acquired using 3D MAPSS with standard parallel imaging (GRAPPA2, Acceleration factor [AF] = 2) with an acquisition time of approximately 12 minutes. The study details are as follows:

1. **Multicenter Orthopaedic Outcomes Network (MOON)[47]:** MRI data were collected from four clinical sites, comprising a total of 236 subjects who were at 10+ years after Anterior Cruciate Ligament Reconstruction (ACLR). Specifically, this study included knees from control subjects and contralateral knees from ACLR patients, with no metal artifacts. Subjects were stratified into 85% training and 15% validation sets.

2. **Corticosteroid Meniscectomy Trial of Extended-Release Triamcinolone after Arthroscopic Partial Meniscectomy (CoMeT)**[48]**:** MRI data were acquired from 105 patients from three sites, resulting in 206 volumes. Like the MOON study, subjects were stratified in 85%-15% split for model training and validation.

3. **Multivendor Multisite Quantitative MRI Study (Multivendor)**[49,50]**:** 20 control subjects underwent bilateral knee MRI (as a subset of the study cohort). A total of 20 volumes were acquired and held out for testing the performance of the models on non-standard quantitative parallel imaging data.

Table 1a. Summary of participants demographics by cohort

| COHORT | SITES (N) | SUBJECTS (N) | FEMALE (N) | AGE (year, MEAN ± SD) | BMI (kg/m$^2$, MEAN ± SD) | SCANS (n) | TRAIN / VAL | TEST |
|---|---|---|---|---|---|---|---|---|
| **MOON** | 4 | 236 | 123 | 33.7 ± 5.1 | 25.7 ± 4.8 | 282 | 246 / 36 | – |
| **COMET** | 3 | 105 | 47 | 55.5 ± 8.2 | 29.6 ± 5.2 | 206 | 173 / 33 | – |
| **MULTIVENDOR** | 1 | 20 | 9 | 40.6 ± 16.7 | 22.8 ± 3.4 | 20 | | 20 |

Table 1b. Scanner and coil details of each site

| Cohort | Site | Scanner | Coil |
|---|---|---|---|
| **MOON** | CCF | Siemens Skyra | 15Rx |
| | OSU | Philips Achieva | 16Rx |
| | VU | Philips Achieva | 16Rx |
| | WashU | Siemens Vida | 18Rx |
| **CoMeT** | CCF | Siemens Skyra | 15Rx |
| | MGB | Siemens Verio | 8Rx |
| | VU | Philips Achieva | 16Rx |
| **Multivendor** | CCF | Siemens Prisma | 15Rx |

CCF: Cleveland Clinic Foundation; OSU: Ohio State University; VU: Vanderbilt University: WashU: University of Washington; MGB: Mass General Brigham

Table 1c. MR protocol parameters

| Parameters | Sagittal 3D DESS WE | Sagittal 3D MAPSS T1ρ/T2 FS |
|---|---|---|
| **Field of view (mm)** | 140 | 140 |
| **Acquisition Matrix** | 384x307x160 | 320x160x24 |
| **Resolution (mm$^3$)** | 0.36x0.36x0.7 | 0.44x0.44x4 |
| **TR/TE (ms)** | 17.55/6.02 | 6.31/2.87 |
| **Acquisition Time (min:s)** | 5:56 | 12:33 |

TR: repetition time; TE: echo time; DESS: dual echo steady state; WE: water excitation. T1ρ imaging spin-lock frequency = 500 Hz with time of spin lock (TSL) = 0/10/30/70 ms; for T2 imaging, preparation TE = 0/20.1/40.1/60.2 ms.

### *B. IMAGE PREPROCESSING AND DATA PREPARATION*

The DESS-based workflow for T1ρ and T2 quantification was used as the ground truth in this study that includes: 1) Rigid registration aligning DESS images and all other echo images to the first echo of the MAPSS; 2) Automatic segmentation of cartilage and meniscus from DESS images using the previously developed Unet-cGAN model[25]; 3) Manual correction of the predicted segmentation masks to create ground truth segmentation masks; 4) Overlay of segmentation mask to the T1ρ and T2 parameter maps for each defined cartilage and meniscal regions. T1ρ and T2 parameter maps were obtained by a voxel-wise two parameter mono-exponential fitting. Voxels with fitted T1ρ > 150 ms, fitted T2 > 130 ms, or goodness-of-fit $R^2 < 0.9$ were excluded during the analysis to remove voxels to minimize potential errors including partial volume averaging effects from adjacent fluids.

For deep learning model training, each slice in echo image volume, corresponding DESS volumes, and segmentation mask volumes were resized to 320 × 320 pixels to ensure uniform input dimensions. Linear and nearest neighbor interpolations were used for image and mask interpolation respectively. Input echo volumes were normalized via min-max scaling and linearly mapped to the range [−1, 1]. Each training sample consisted of source echo images (4x320x320, 4 channels for the first 4 echoes), target DESS (1x320x320), and ground truth mask (7x320x320, 7 channels for femoral cartilage (FC), lateral tibial cartilage (LTC), medial tibial cartilage (MTC), patellar cartilage (PAT), lateral meniscus (LM), medial meniscus (MM), and background).

### *C. MODEL DESCRIPTION*

We developed a multi-task conditional generative adversarial network (MT-cGAN) to jointly perform synthetic DESS image generation and cartilage and meniscus segmentation from MAPSS echo images. The generator follows a U-Net[51] architecture with a shared encoder and two parallel decoders. The shared encoder consists of stacked convolutional layers that capture low-level and high-level features common to both tasks, reducing the total number of parameters while promoting generalization across domains. One decoder reconstructs synthetic DESS images using transposed convolutions and skip connections to preserve spatial detail, while the other predicts segmentation masks for six cartilage and meniscus compartments. This dual-decoder design prevents task interference, a common issue in multi-task learning, by allowing each decoder to optimize its output without conflicting gradients. Therefore, the model can be trained in three modes: 1) generation mode where synthetic DESS images are generated from echo images, 2) segmentation mode where cartilage and meniscus segmentation are generated from echo images, and 3) multitask mode where the model optimizes both tasks simultaneously. Skip connections between encoder and decoder layers at corresponding resolutions preserve spatial details and improve gradient flow during training. The discriminator receives a concatenation of the source echo images with either the generated output (synthetic DESS image, segmentation mask, or both) or the corresponding real counterpart, and learns to distinguish between real and synthetic pairs. We used instance normalization and LeakyReLU activations to maintain the training stability. The generator loss is a weighted combination of adversarial binary cross-entropy loss (encourages realism), L1 loss (for creating synthetic DESS), Dice loss and multi-class cross-entropy (for improved segmentation). Loss weights are configurable to balance the contribution of each component based on the task (generation, segmentation, or multitask). The training process alternates between updating the generator and discriminator, with configurable step ratios to stabilize convergence.

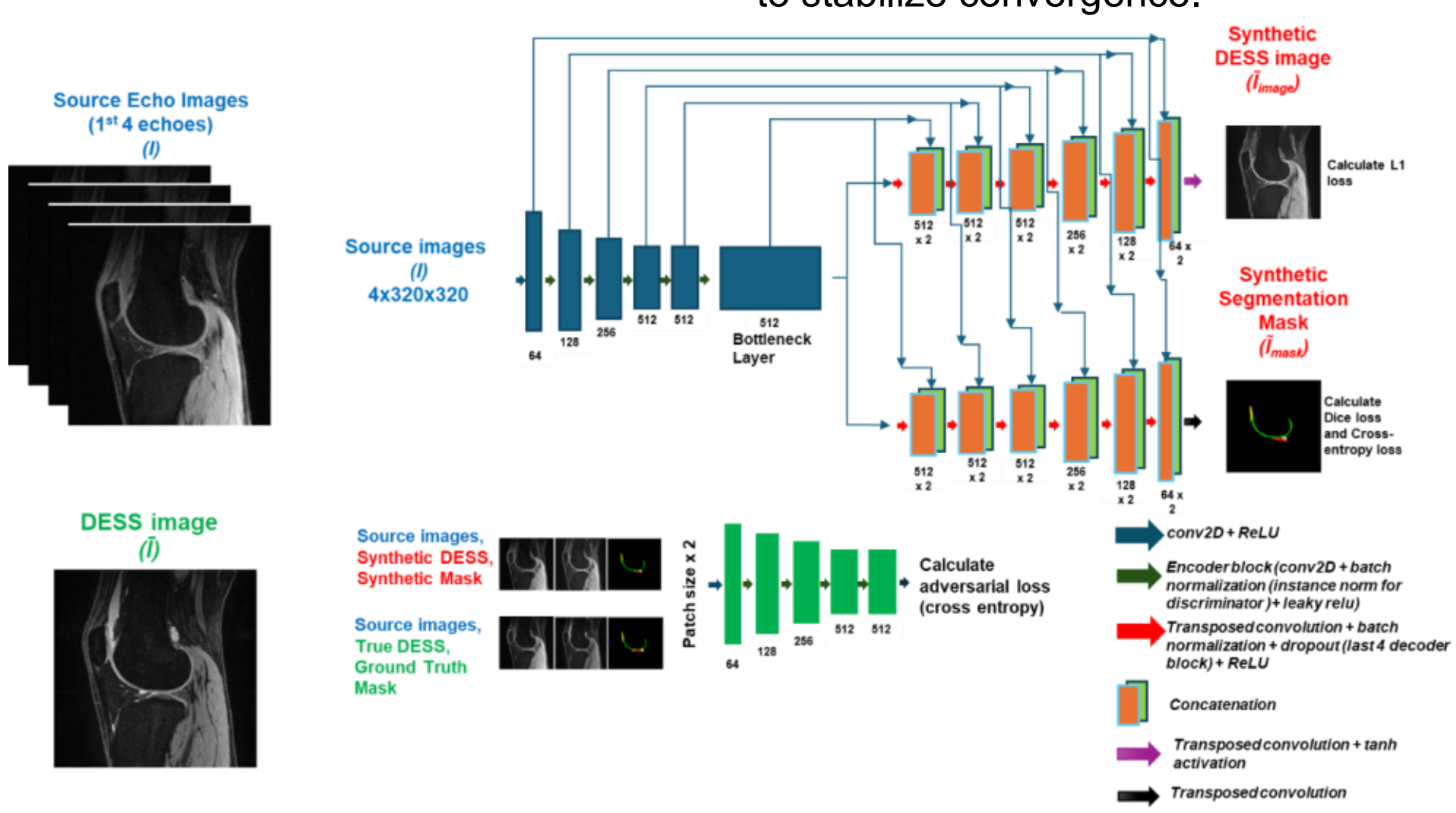

**Figure 1:** MT-cGAN architecture for multitask learning of DESS synthesis and cartilage and meniscus segmentation

The generator was optimized using a composite loss function whose components were weighted based on task relevance. For multitask training, the total generator loss included: adversarial binary cross-entropy loss ($\alpha_{adv}$=1),

L1 loss for synthetic DESS reconstruction ($\alpha_{L1}$=100), Dice loss for segmentation accuracy ($\alpha_{Dice}$=100), and multi-class cross-entropy loss ($\alpha_{CE}$=50). These loss weights were selected based on a grid search over candidate values, optimizing for validation performance on segmentation accuracy. The discriminator was trained to distinguish between real and generated outputs by minimizing binary cross-entropy loss with label smoothing (real label = 0.9, fake = 0.0). Its input consisted of a concatenation of the source MAPSS echo images with either the real or generated DESS image and/or segmentation mask, depending on the training mode. Training was done over 100 epochs using the Adam optimizer with a learning rate of $1\times10^{-5}$. Model validation was performed after every epoch using volume-level Dice score, and the best performing model in the validation set was used for held-out test set evaluation.

### *D. EXPERIMENTAL SETUP AND EVALUATION*

To evaluate the effectiveness of our proposed multi-task model, we compared it against four alternative methods that aim to segment cartilage and meniscus, and enable T1ρ and T2 quantification, without relying on DESS images. All methods used MAPSS echo images as input and were benchmarked against the ground truth (manual correction of DESS-based segmentation with Unet-cGAN[25]). The evaluated methods are described below:

1. **MAPSS-NTL:** Unet-cGAN[25], originally trained on DESS images for segmentation, was directly applied on the MAPSS first echo image without additional training,
2. **MAPSS-TL:** Transfer learning (TL) was performed on Unet-cGAN using the first echo of MAPSS and segmentation from registered DESS as 'ground truth'. The TL model was then applied to test cases of MAPSS echo images.
3. **Synthetic DESS (MT-cGAN$_{Gen}$) + Segmentation (Unet-cGAN):** A two-stage modular approach. The first stage synthesized a DESS-like image from MAPSS echo images using MT-cGAN in generation mode. In the second stage, the synthetic DESS was segmented using the pretrained Unet-cGAN model. This method tested whether a DESS proxy can enable reuse of the existing DESS-based segmentation model.
4. **Segmentation (MT-cGAN$_{Seg}$):** Training proposed MT-cGAN in segmentation mode from scratch on MAPSS echo images. It served as a single-task baseline for direct segmentation without image synthesis.
5. **Multitask (MT-cGAN)** (**Proposed Model**): Our proposed method jointly performed synthetic DESS generation and cartilage/meniscus segmentation in a single network. It employed a shared encoder with two decoders: one for image synthesis and one for segmentation, allowing for shared feature learning and task-specific optimization. This approach was designed to improve anatomical consistency and segmentation accuracy simultaneously.

All models were evaluated across six anatomical regions (FC, LTC, MTC, PAT, LM, MM). Performance was assessed in terms of both segmentation accuracy and the resulting compartment wise T1ρ and T2 quantification, using the DESS-based workflow as ground truth. Segmentation accuracy was assessed based on Dice score, coefficients of variation (CVs), and Bland-Altman analysis were used to assess T1ρ and T2 quantification. Statistical significance among methods was assessed via Wilcoxon signed-rank test.

## III. RESULTS

Figure 2 shows the ground truth annotations, predicted segmentation outputs, and error maps for all methods on selected slices from a representative test volume. We observed that MT-cGAN consistently produces segmentations that closely match the ground truth, with minimal over- or under-segmentation, particularly in challenging regions such as femoral cartilage, and medial meniscus. Without transfer learning, MAPSS-NTL (Unet-cGAN originally trained on DESS images) performed poorly on echo images due to the domain shift. Transfer learning (MAPSS-TL) adapts to that domain shift and performs well on the echo images. However, we observed that MAPSS-TL missed the anterior and posterior horns of the medial meniscus. MT-cGAN$_{Gen}$ produces synthetic DESS images but the generation quality impacts the subsequent segmentation using Unet-cGAN. We observed that this method provides highly precise segmentation. However, it consistently under-segments (high precision, low recall). The segmentation only mode, MT-cGAN$_{Seg}$ performed similarly to MAPSS-TL.

Segmentation performance of the proposed MT-cGAN in comparison to other methods in the test dataset are shown in Table 2. Boxplots of Dice scores are shown in Figure 3. Both MAPSS-NTL and MAPSS-TL used the first echo from MAPSS images for segmentation. Proposed approach (MT-cGAN) achieved the highest Dice score across all classes [0.80-0.86], when compared to all the other methods. Wilcoxon signed rank test showed that MT-cGAN (mean Dice = 0.8396) outperforms MAPSS-TL (mean Dice = 0.8152, state-of-the-art Unet-cGAN model with transfer learning) significantly ($p<10^{-5}$). The segmentation only mode MT-cGAN$_{Seg}$ (mean Dice = 0.8059) performs similarly to MAPSS-TL and there was no statistical difference (p=0.052). Statistical significance per class for each method, comparison of precision and recall are shown in Supplemental Figure 1 and Supplemental Tables 1 and 2.

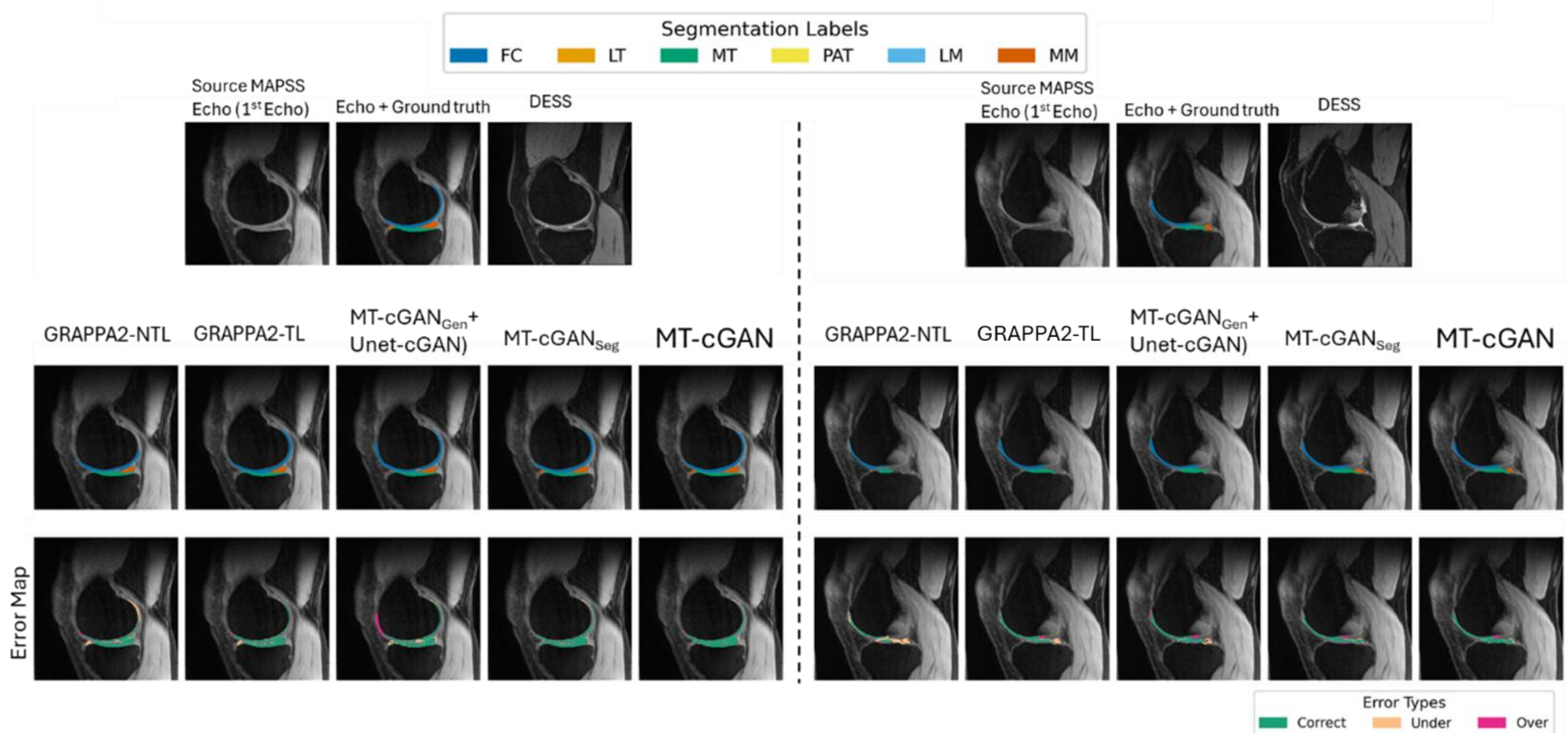

Figure 2: Representative example of segmentation predictions by different methods. Source echo images (with only the first echo displayed), echo overlaid with ground truth annotation (color-coded) and corresponding DESS images are in top row. Middle row shows prediction from different methods. Bottom row shows error maps containing accurate prediction, over- and under-segmentation by each method (color-coded).

Table 2: Comparison of Dice scores [Mean (SD)] in cartilage and meniscus segmentation

| **Model** | **Femoral Cartilage** | **Lateral Tibial Cartilage** | **Medial Tibial Cartilage** | **Patellar Cartilage** | **Lateral Meniscus** | **Medial Meniscus** | **Mean** |
|---|---|---|---|---|---|---|---|
| **MAPSS-NTL** | 0.6266 (0.1613) | 0.6779 (0.1793) | 0.6787 (0.1372) | 0.3608 (0.2041) | 0.5687 (0.1802) | 0.5117 (0.1657) | 0.5707 (0.1713) |
| **MAPSS-TL** | 0.8195 (0.0554) | 0.8468 (0.0387) | 0.8240 (0.0505) | 0.8033 (0.0424) | 0.8038 (0.0527) | 0.7941 (0.0602) | 0.8152 (0.0500) |
| **Synthetic DESS (MT-cGAN$_{Gen}$) + Segmentation (Unet-cGAN)** | 0.7439 (0.0730) | 0.6648 (0.1101) | 0.6850 (0.0995) | 0.6552 (0.0976) | 0.6583 (0.1086) | 0.6218 (0.1173) | 0.6715 (0.1010) |
| **Segmentation (MT-cGAN$_{Seg}$)** | 0.8042 (0.0440) | 0.8247 (0.0682) | 0.8270 (0.0374) | 0.7877 (0.0504) | 0.7908 (0.0632) | 0.8013 (0.0593) | 0.8059 (0.0537) |
| **Multitask (MT-cGAN)** | 0.8404 (0.0572) | 0.8639 (0.0497) | 0.8548 (0.0652) | 0.8006 (0.0857) | 0.8466 (0.0689) | 0.8313 (0.0729) | 0.8396 (0.0666) |

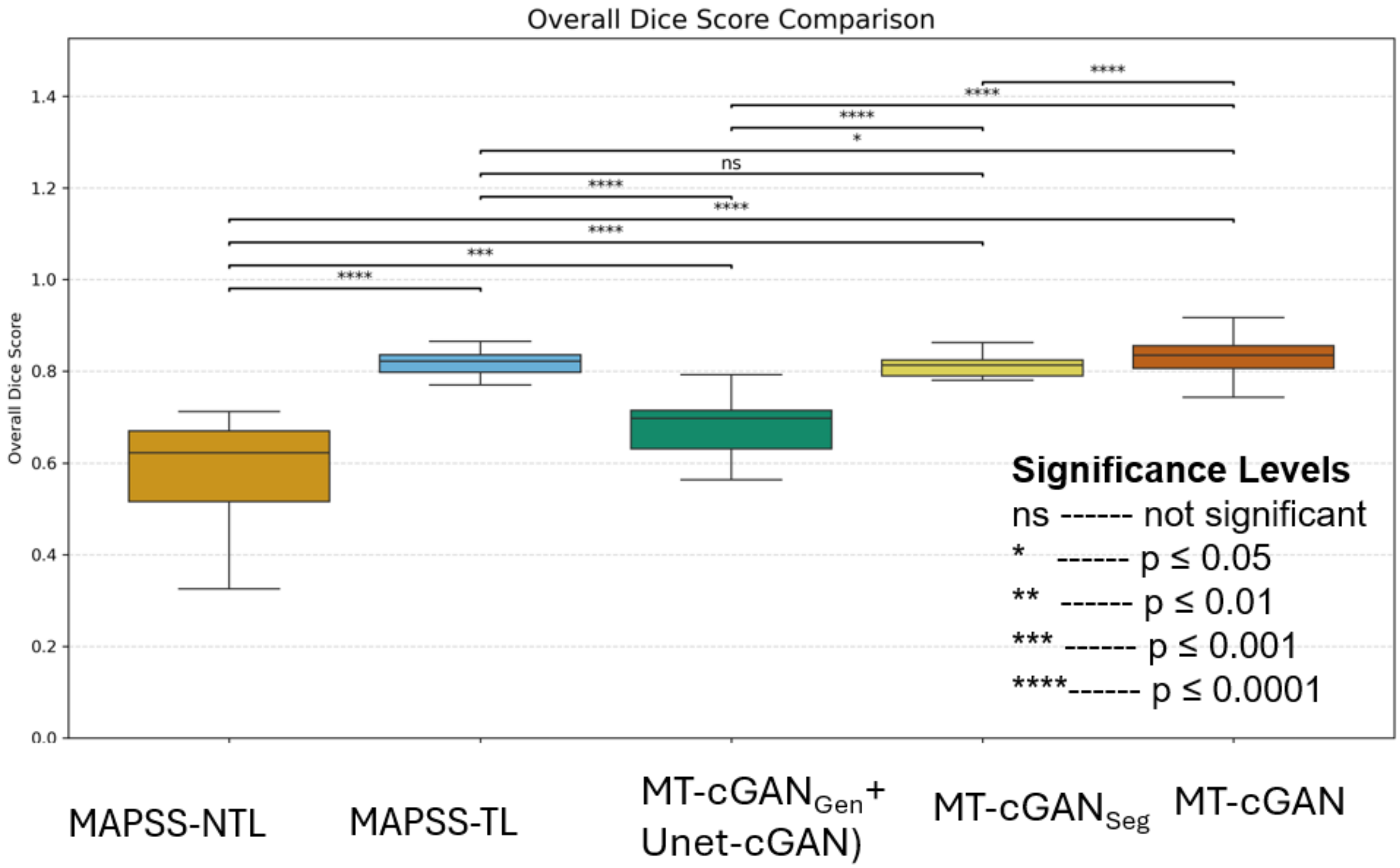

Figure 3: Dice scores comparison among different methods for cartilage and meniscus segmentation. Each box represents the distribution of Dice scores across 20 test cases. Statistical significance was assessed using the Wilcoxon signed-rank test, with annotations showing p-values as asterisks.

Table 3: Mean coefficients of variation of T1ρ and T2 from different models compared with the reference DESS-based protocol

| Cartilage Segmentation Methods | Map -ping | Mapping obtained from | Femoral Cartilage | Lateral Tibial Cartilage | Medial Tibial Cartilage | Patellar Cartilage | Lateral Meniscus | Medial Meniscus | Mean |
|---|---|---|---|---|---|---|---|---|---|
| MAPSS-NTL | | | 1.30% | 3.50% | 2.33% | 5.45% | 8.71% | 4.97% | 4.38% |
| MAPSS-TL | | | 1.30% | 2.44% | 2.81% | 3.45% | 2.68% | 2.70% | 2.56% |
| Synthetic DESS (MT-cGAN$_{Gen}$) + Segmentation (Unet-cGAN) | $T_{1\rho}$ | Reference | 1.18% | 3.13% | 1.70% | 3.81% | 3.52% | 2.02% | 2.56% |
| Segmentation (MT-cGAN$_{Seg}$) | | | 1.08% | 1.99% | 2.70% | 2.83% | 3.72% | 2.89% | 2.53% |
| Multitask (MT-cGAN) | | | **0.74%** | **1.29%** | **1.51%** | **3.01%** | **2.15%** | **2.35%** | **1.84%** |
| MAPSS-NTL | | | 0.97% | 3.98% | 1.99% | 5.57% | 8.14% | 6.25% | 4.49% |
| MAPSS-TL | | | 1.37% | 2.20% | 2.22% | 3.99% | 2.99% | 2.67% | 2.57% |
| Synthetic DESS (MT-cGAN$_{Gen}$) + Segmentation (Unet-cGAN) | $T_2$ | Reference | 1.06% | 3.40% | 1.57% | 4.11% | 3.09% | 2.28% | 2.58% |
| Segmentation (MT-cGAN$_{Seg}$) | | | 1.12% | 1.72% | 2.19% | 3.17% | 3.14% | 2.88% | 2.37% |
| Multitask (MT-cGAN) | | | **0.63%** | **1.16%** | **1.26%** | **3.32%** | **2.09%** | **2.38%** | **1.81%** |

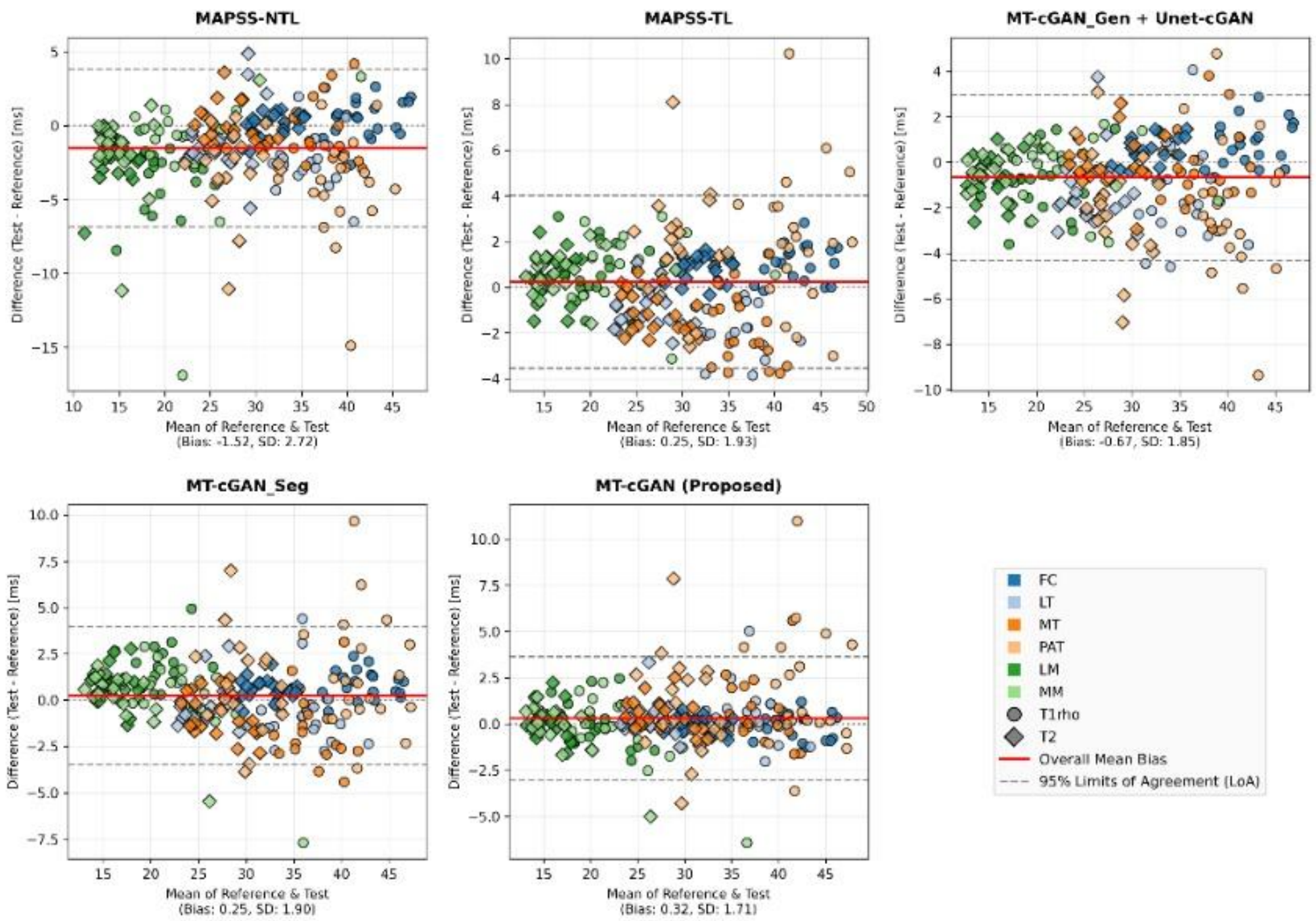

Figure 4: Quantitative measurement agreement and error distribution across T1ρ or T2 pipelines. Combined Bland-Altman plots show measurement agreement against the DESS-based reference standard across all 20 subjects and anatomical sub-compartments.

Table 3 and Figure 4 show the evaluation of the T1ρ or T2 quantification, detailing compartment-wise CVs and Bland-Altman plots of agreement for different segmentation approaches, compared to the reference DESS-based protocol. For both T1ρ and T2, the proposed Multitask model (MT-cGAN) consistently achieved the lowest mean CV values: 1.84% (T1ρ) and 1.81% (T2), indicating the most consistent mapping among all tested methods. It also achieved small bias of +0.32ms (95% Limits of Agreement [LoA]: [-3.03, 3.66]ms). This suggests that the multitask approach not only produces accurate segmentations but also provides robust MRI quantification across cartilage and meniscus compartments. Compared to the established workflow with DESS segmentation, MAPSS-NTL showed the highest variability (global CVs > 4%) and a significant global negative bias of -1.52ms (95% LoA: [-6.86, 3.82]ms). Transfer learning improved the CVs (MAPSS-TL CV <3%) and diminished the bias to +0.25 ms (95% LoA: [-3.54, 4.03]ms). MT-cGAN$_{Gen}$ + Unet-cGAN method also performed competitively, though with slightly higher variability and bias (-0.67ms) in certain compartments (e.g., lateral tibial cartilage).

## IV. DISCUSSION

This study shows that the proposed Multitask Conditional GAN (MT-cGAN) can reliably segment knee cartilage and menisci directly from T1ρ and T2 echo images while simultaneously generating a DESS-like image. The proposed MT-cGAN outperformed the state-of-the-art conditional GAN with transfer learning and produced the most consistent T1ρ and T2 relaxometry measurements relative to the established DESS-based workflow. These findings highlight the feasibility of eliminating high resolution morphological scans (i.e., DESS), thereby streamlining quantitative MRI protocols and improving their clinical utility.

Segmentation from echo images alone is feasible, as demonstrated by both MT-cGAN$_{Seg}$ and MAPSS-TL (transfer learned Unet-cGAN). When comparing MAPSS-TL vs MAPSS-NTL, transfer learning provided segmentation accuracy comparable to DESS-based segmentation due to learning the domain-specific representations of the echo images. MAPSS-NTL and MAPSS-TL workflows use only the first echo for segmentation. When no training was performed (MAPSS-NTL), several regions were not classified at all, leading to low Dice scores. Figure 3 shows that transfer learning improved Dice scores across all compartments of cartilage and menisci with mean ΔDice = 24.45%. After transfer learning, we observed smaller standard deviations in MAPSS-TL workflow as well, implying more robust and consistent performance. MT-cGAN$_{Seg}$ performs similarly to the MAPSS-TL workflow (mean Dice 0.8059 vs. 0.8152; $p = 0.052$, Wilcoxon signed-rank test). While MT-cGAN$_{Seg}$ uses additional echo images as inputs, it does not include the feature matching loss which penalizes differences in intermediate discriminator features between real and predicted masks, unlike MAPSS-TL. This shows that echo images contain sufficient information for segmentation. However, the segmentation models had to be explicitly trained to learn echo image contrast. With improved segmentation performance, T1ρ and T2 quantification based on segmentation using MAPSS echo images also provided high accuracy relative to DESS-based workflow (global CV ~3%) for both MAPSS-TL and MT-cGAN$_{Seg}$. The two-stage approach, segmentation following synthesis (MT-cGAN$_{Gen}$+Unet-cGAN), did not perform adequately. The image generation quality of MT-cGAN$_{Gen}$ was not sufficient to support segmentation using Unet-cGAN. This is most likely due to subtle differences in synthetic DESS images that introduced a domain mismatch for Unet-cGAN, while Unet-cGAN was trained on real DESS images. Interestingly, we observed that the lower segmentation performance did not result in higher CVs of T1ρ and T2 quantification, due to the workflow's conservative segmentation (high precision, low recall). While single-task models adapted to the echo domain (either via transfer learning or training from scratch) and achieved reasonable accuracy, their performance plateaued due to the limited contrast information available in the echo images.

The proposed multitask model, MT-cGAN, overcomes the limits of single-task segmentation models by introducing structural supervision through an image synthesis task. MT-cGAN outperformed single-task models, i.e., MAPSS-TL and MT-cGAN$_{Seg}$ across all cartilage and menisci compartments. The shared encoder, jointly optimized for DESS synthesis and segmentation, learns to extract DESS-like structural features, i.e., sharper cartilage and menisci boundaries, that are not explicitly present in the echo images alone. This bridges the domain gap that degraded segmentation performance in prior work[22]. This resulted in a significant improvement in the image segmentation task over single-task models (mean ΔDice = 3.37%; $p<0.001$, Wilcoxon signed-rank test). MT-cGAN achieved a mean Dice score of 0.8396 across all six structures, demonstrating that the auxiliary DESS generation task provides a strong inductive bias that compensates for the lack of cartilage/meniscus details in echo images. The high segmentation performance directly improved T1ρ and T2 quantification, outperforming both MAPSS-TL and MT-cGAN$_{Seg}$, and achieving the lowest CVs, below 2%, for both T1ρ and T2 measurements.

One limitation of this study is that ground truth segmentation was created using DESS images. Manual segmentation directly from echo images could be used as ground truth; however, this is challenging due to the low resolution and blurred edges, which introduces significant intra- and inter-reader variations. DESS has better cartilage tissue contrasts compared to other imaging sequences, making it preferred for segmentation purpose.[52,53] The only caveat is registration between DESS and echo images. Distortion correction is commonly used for clinical imaging to reduce image distortion caused by field inhomogeneity or gradient imperfections. This correction was applied to DESS but not on MAPSS images, and when this nonlinear distortion correction was applied in a bigger scale, it created misregistration that were not corrected with the registration pipeline. Another limitation is that our held-out test set only included control knees. Although this minimized ambiguity in segmentation and relaxometry measurements, further testing is needed in patients with OA, postoperative changes, or other pathology. Future work will address these limitations and possibly explore segmentation and quantification from accelerated MAPSS imaging, which would enable reliable quantitative assessment in under 3 minutes[49].

## V. CONCLUSIONS

This study demonstrates the feasibility of automatic and reliable quantification of T1ρ and T2 relaxation times of cartilage and menisci without the need for DESS or other high resolution morphological images by developing a novel multitask deep learning model. Eliminating the DESS sequence reduces scan time by approximately 6 minutes, improves patient comfort, reduces the risk of motion artifacts, and removes the need of registration, making the workflow more practical for routine clinical use. Such efficiency gains can greatly facilitate implementation of quantitative MRI, supporting earlier diagnosis and improved prognosis of significant musculoskeletal disorders such as osteoarthritis.

## REFERENCES

1. Foster, A. L. *et al.* Distribution of Arthritis Subtypes Among Adults With Arthritis in the United States, 2017–March 2020. *Prev. Chronic. Dis.* **22**, E28 (2025).
2. A National Public Health Agenda for Osteoarthritis: 2020 Update.
3. Eck, B. L. *et al.* Quantitative MRI for Evaluation of Musculoskeletal Disease: Cartilage and Muscle Composition, Joint Inflammation, and Biomechanics in Osteoarthritis. *Invest. Radiol.* **58**, 60–75 (2023).
4. Li, X. *et al.* Cartilage compositional MRI—a narrative review of technical development and clinical applications over the past three decades. *Skeletal Radiol.* **53**, 1761–1781 (2024).
5. Li, X. & Winalski, C. S. Chapter 8 - Relaxometry: Applications in Musculoskeletal Systems. in *Quantitative Magnetic Resonance Imaging* (eds Seiberlich, N. et al.) vol. 1 185–214 (Academic Press, 2020).
6. Wirth, W. *et al.* Clinical validation of fully automated cartilage transverse relaxation time (T2) and thickness analysis using quantitative DESS magnetic resonance imaging. *Magn. Reson. Mater. Phys. Biol. Med.* 1–13 (2025).
7. Chaudhari, A. S. *et al.* Five-minute knee MRI for simultaneous morphometry and T2 relaxometry of cartilage and meniscus and for semiquantitative radiological assessment using double-echo in steady-state at 3T. *J. Magn. Reson. Imaging* **47**, 1328–1341 (2018).
8. Fripp, J., Crozier, S., Warfield, S. K. & Ourselin, S. Automatic segmentation and quantitative analysis of the articular cartilages from magnetic resonance images of the knee. *IEEE Trans. Med. Imaging* **29**, 55–64 (2009).
9. Tamez-Peña, J. *et al.* Atlas based method for the automated segmentation and quantification of knee features: Data from the osteoarthritis initiative. in *2011 IEEE International Symposium on Biomedical Imaging: From Nano to Macro* 1484–1487 (IEEE, 2011).
10. Shan, L., Zach, C., Charles, C. & Niethammer, M. Automatic atlas-based three-label cartilage segmentation from MR knee images. *Med. Image Anal.* **18**, 1233–1246 (2014).
11. Lee, J.-G., Gumus, S., Moon, C. H., Kwoh, C. K. & Bae, K. T. Fully automated segmentation of cartilage from the MR images of knee using a multi-atlas and local structural analysis method. *Med. Phys.* **41**, 092303 (2014).
12. Zarychta, P. Atlas-based segmentation in extraction of knee joint bone structures from CT and MR. *Sensors* **22**, 8960 (2022).
13. Solloway, S., Hutchinson, C. E., Waterton, J. C. & Taylor, C. J. The use of active shape models for making thickness measurements of articular cartilage from MR images. *Magn. Reson. Med.* **37**, 943–952 (1997).
14. Fripp, J., Crozier, S., Warfield, S. & Ourselin, S. Automatic initialization of 3D deformable models for cartilage segmentation. in *Digital Image Computing: Techniques and Applications (DICTA'05)* 74–74 (IEEE, 2005).
15. Vincent, G. Fully automatic segmentation of the knee joint using active appearance models. in *Proc. of Medical Image Analysis for the Clinic: A Grand Challenge, MICCAI workshop., Beijing, Sept. 2010* 224–230 (2010).
16. Tamez-Pena, J. G. *et al.* Unsupervised segmentation and quantification of anatomical knee features: data from the Osteoarthritis Initiative. *IEEE Trans. Biomed. Eng.* **59**, 1177–1186 (2012).
17. González, G. & Escalante-Ramírez, B. Knee cartilage segmentation using active shape models and local binary patterns. in *Optics, Photonics, and Digital Technologies for Multimedia Applications III* vol. 9138 159–169 (SPIE, 2014).
18. Prasoon, A. *et al.* Deep feature learning for knee cartilage segmentation using a triplanar convolutional neural network. in *International conference on medical image computing and computer-assisted intervention* 246–253 (Springer, 2013).
19. Liu, F. *et al.* Deep convolutional neural network and 3D deformable approach for tissue segmentation in musculoskeletal magnetic resonance imaging. *Magn. Reson. Med.* **79**, 2379–2391 (2018).
20. Zhou, Z., Zhao, G., Kijowski, R. & Liu, F. Deep convolutional neural network for segmentation of knee joint anatomy. *Magn. Reson. Med.* **80**, 2759–2770 (2018).
21. Raj, A., Vishwanathan, S., Ajani, B., Krishnan, K. & Agarwal, H. Automatic knee cartilage segmentation using fully volumetric convolutional neural networks for evaluation of osteoarthritis. in *2018 IEEE 15th International Symposium on Biomedical Imaging (ISBI 2018)* 851–854 (IEEE, 2018).
22. Norman, B., Pedoia, V. & Majumdar, S. Use of 2D U-Net convolutional neural networks for automated cartilage and meniscus segmentation of knee MR imaging data to determine relaxometry and morphometry. *Radiology* **288**, 177–185 (2018).
23. Ambellan, F., Tack, A., Ehlke, M. & Zachow, S. Automated segmentation of knee bone and cartilage combining statistical shape knowledge and convolutional neural networks: Data from the Osteoarthritis Initiative. *Med. Image Anal.* **52**, 109–118 (2019).
24. Kessler, D. A. *et al.* The optimisation of deep neural networks for segmenting multiple knee joint tissues from MRIs. *Comput. Med. Imaging Graph.* **86**, 101793 (2020).
25. Gaj, S., Yang, M., Nakamura, K. & Li, X. Automated cartilage and meniscus segmentation of knee MRI with conditional generative adversarial networks. *Magn. Reson. Med.* **84**, 437–449 (2020).
26. Wirth, W. *et al.* Accuracy and longitudinal reproducibility of quantitative femorotibial cartilage measures derived from automated U-Net-based segmentation of two different MRI contrasts: data from the osteoarthritis initiative healthy reference cohort. *Magn. Reson. Mater. Phys. Biol. Med.* **34**, 337–354 (2021).
27. Eckstein, F. *et al.* Detection of differences in longitudinal cartilage thickness loss using a deep-learning automated segmentation algorithm: Data from the Foundation for the National Institutes of Health biomarkers study of the osteoarthritis initiative. *Arthritis Care Res.* **74**, 929–936 (2022).
28. Panfilov, E., Tiulpin, A., Nieminen, M. T., Saarakkala, S. & Casula, V. Deep learning-based segmentation of knee MRI for fully automatic subregional morphological assessment of cartilage tissues: data from the osteoarthritis initiative. *J. Orthop. Res.* **40**, 1113–1124 (2022).
29. Yang, M. *et al.* Automated knee cartilage segmentation for heterogeneous clinical MRI using generative adversarial networks with transfer learning. *Quant. Imaging Med. Surg.* **12**, 2620–2633 (2022).
30. Chen, H. *et al.* Knee bone and cartilage segmentation based on a 3D deep neural network using adversarial loss for prior shape constraint. *Front. Med.* **9**, 792900 (2022).
31. Kim-Wang, S. Y. *et al.* Auto-segmentation of the tibia and femur from knee MR images via deep learning and its application to cartilage strain and recovery. *J. Biomech.* **149**, 111473 (2023).
32. Schmidt, A. M. *et al.* Generalizability of deep learning segmentation algorithms for automated assessment of cartilage morphology and MRI relaxometry. *J. Magn. Reson. Imaging* **57**, 1029–1039 (2023).
33. Liu, H., Sun, Y., Cheng, X. & Jiang, D. Prior-based 3D U-Net: A model for knee-cartilage segmentation in MRI images. *Comput. Graph.* **115**, 167–180 (2023).
34. Martel-Pelletier, J., Paiement, P. & Pelletier, J.-P. Magnetic resonance imaging assessments for knee segmentation and their use in combination with machine/deep learning as predictors of early osteoarthritis diagnosis and prognosis. *Ther. Adv. Musculoskelet. Dis.* **15**, 1759720X231165560 (2023).
35. Sengar, S. S. *et al.* Multi-planar 3D knee MRI segmentation via UNet inspired architectures. *Int. J. Imaging Syst. Technol.* **33**, 985–998 (2023).
36. Kulseng, C. P. S. *et al.* Automatic segmentation of human knee anatomy by a convolutional neural network applying a 3D MRI protocol. *BMC Musculoskelet. Disord.* **24**, 41 (2023).
37. Guo, J. *et al.* Automated measurement and grading of knee cartilage thickness: a deep learning-based approach. *Front. Med.* **11**, 1337993 (2024).
38. Ferreira, D. L. *et al.* A memory based model for cartilage and meniscus segmentation in 3D knee MRI. *Sci. Rep.* (2025).
39. Moorthy, C., Shafeek, A., Gurunathan, V. & Sudhakar, R. 3D CSFA-UNet: a unified attention-driven deep learning framework for accurate knee MRI segmentation and osteoarthritis severity classification. *Sci. Rep.* (2026).
40. Neubert, A. *et al.* Simultaneous super-resolution and contrast synthesis of routine clinical magnetic resonance images of the knee for improving automatic segmentation of joint cartilage: data from the Osteoarthritis Initiative. *Med. Phys.* **47**, 4939–4948 (2020).
41. Caruana, R. Multitask learning. *Mach. Learn.* **28**, 41–75 (1997).
42. Sener, O. & Koltun, V. Multi-task learning as multi-objective optimization. *Adv. Neural Inf. Process. Syst.* **31**, (2018).
43. Vandenhende, S. *et al.* Multi-task learning for dense prediction tasks: A survey. *IEEE Trans. Pattern Anal. Mach. Intell.* **44**, 3614–3633 (2021).
44. Moeskops, P. *et al.* Deep learning for multi-task medical image segmentation in multiple modalities. in *International conference on medical image computing and computer-assisted intervention* 478–486 (Springer, 2016).
45. Zhou, Y. *et al.* Multi-task learning for segmentation and classification of tumors in 3D automated breast ultrasound images. *Med. Image Anal.* **70**, 101918 (2021).
46. Karkalousos, D., Isgum, I., Marquering, H. & Caan, M. W. MultiTask learning for accelerated-MRI reconstruction and segmentation of brain lesions in multiple sclerosis. in *Medical Imaging with Deep Learning* 991–1005 (PMLR, 2024).
47. Xie, D. *et al.* Multi-vendor multi-site quantitative MRI analysis of cartilage degeneration 10 Years after anterior cruciate ligament reconstruction: MOON-MRI protocol and preliminary results. *Osteoarthritis Cartilage* **30**, 1647–1657 (2022).
48. Sullivan, J. K. *et al.* The Corticosteroid Meniscectomy Trial of Extended-Release Triamcinolone Injection After Arthroscopic Partial Meniscectomy: Protocol for a Double-Blind Randomized Controlled Trial. *Orthop. J. Sports Med.* **11**, 23259671231150812 (2023).
49. Kim, J. *et al.* Highly Accelerated T1ρ Imaging in 3 min: Comparison Between Compressed Sensing and Deep Learning Reconstruction. *NMR Biomed.* **39**, e70226 (2026).
50. Kim, J. *et al.* Multi-vendor multi-site T1ρ and T2 quantification of knee cartilage. *Osteoarthritis Cartilage* **28**, 1539–1550 (2020).
51. Ronneberger, O., Fischer, P. & Brox, T. U-net: Convolutional networks for biomedical image segmentation. *Med. Image Comput. Comput.-Assist. Interv.* **9351**, 234–241 (2015).
52. Eckstein, F. *et al.* Two year longitudinal change and test-retest-precision of knee cartilage morphology in a pilot study for the osteoarthritis initiative. *Osteoarthritis Cartilage* **15**, 1326–1332 (2007).
53. Peterfy, C. G., Schneider, E. & Nevitt, M. The osteoarthritis initiative: report on the design rationale for the magnetic resonance imaging protocol for the knee. *Osteoarthritis Cartilage* **16**, 1433–1441 (2008).